\documentclass{wscpaperproc}
\usepackage{latexsym}
\usepackage{graphicx}
\usepackage{mathptmx}

\usepackage{amsmath}
\usepackage{amsfonts}
\usepackage{amssymb}
\usepackage{amsbsy}
\usepackage{amsthm}

\usepackage{url}
\usepackage{graphicx}
\usepackage{booktabs}
\usepackage{array}
\usepackage{tabularx}
\usepackage{bm}
\usepackage{lscape}
\usepackage{comment}
\usepackage{enumitem}
\graphicspath{{figures/}}
\usepackage{amsmath}
\usepackage{lipsum}

\newcolumntype{s}{>{\hsize=0.6cm}l}
\usepackage{epstopdf}
\usepackage[caption=false]{subfig}
\usepackage{fontawesome}

\usepackage[pdftex,colorlinks=true,urlcolor=blue,citecolor=black,anchorcolor=black,linkcolor=black]{hyperref}

\newtheoremstyle{wsc}
{3pt}
{3pt}
{}
{}
{\bf}
{}
{.5em}
{}

\theoremstyle{wsc}

\begin{document}

%
%

\pagestyle{fancyplain}

\thispagestyle{plain}
\firstPageHead{}

\chead{\fancyplain{}{\itshape Dantas, Costa, Medeiros, Geraldo, Maximo, and Yoneyama}}

\rhead{}
\cfoot{}
\renewcommand{\headrulewidth}{0pt} 

\makeatletter
\let\@internalcite\cite
\def\cite{\def\@citeseppen{-1000}%
    \def\@cite##1##2{(##1\if@tempswa , ##2\fi)}%
    \def\citeauthoryear##1##2##3{##1 ##3}\@internalcite}
\def\citeNP{\def\@citeseppen{-1000}%
    \def\@cite##1##2{##1\if@tempswa , ##2\fi}%
    \def\citeauthoryear##1##2##3{##1 ##3}\@internalcite}
\def\citeN{\def\@citeseppen{-1000}%
    \def\@cite##1##2{##1\if@tempswa, ##2)\else{}\fi}%
    \def\citeauthoryear##1##2##3{##1 (##3)}\@citedata}
\def\citeA{\def\@citeseppen{-1000}%
    \def\@cite##1##2{(##1\if@tempswa , ##2\fi)}%
    \def\citeauthoryear##1##2##3{##1}\@internalcite}
\def\citeANP{\def\@citeseppen{-1000}%
    \def\@cite##1##2{##1\if@tempswa , ##2\fi}%
    \def\citeauthoryear##1##2##3{##1}\@internalcite}
\def\shortcite{\def\@citeseppen{-1000}%
    \def\@cite##1##2{(##1\if@tempswa , ##2\fi)}%
    \def\citeauthoryear##1##2##3{##2 ##3}\@internalcite}
\def\shortciteNP{\def\@citeseppen{-1000}%
    \def\@cite##1##2{##1\if@tempswa , ##2\fi}%
    \def\citeauthoryear##1##2##3{##2 ##3}\@internalcite}
\def\shortciteN{\def\@citeseppen{-1000}%
    \def\@cite##1##2{##1\if@tempswa, ##2\else{}\fi}%
    \def\citeauthoryear##1##2##3{##2 (##3)}\@citedata}
\def\shortciteA{\def\@citeseppen{-1000}%
    \def\@cite##1##2{(##1\if@tempswa , ##2\fi)}%
    \def\citeauthoryear##1##2##3{##2}\@internalcite}
\def\shortciteANP{\def\@citeseppen{-1000}%
    \def\@cite##1##2{##1\if@tempswa , ##2\fi}%
    \def\citeauthoryear##1##2##3{##2}\@internalcite}
\def\citeyear{\def\@citeseppen{-1000}%
    \def\@cite##1##2{(##1\if@tempswa , ##2\fi)}%
    \def\citeauthoryear##1##2##3{##3}\@citedata}
\def\citeyearNP{\def\@citeseppen{-1000}%
    \def\@cite##1##2{##1\if@tempswa , ##2\fi}%
    \def\citeauthoryear##1##2##3{##3}\@citedata}
%
%
%
\def\@citedata{%
    \@ifnextchar [{\@tempswatrue\@citedatax}%
                  {\@tempswafalse\@citedatax[]}%
}

\def\@citedatax[#1]#2{%
\if@filesw\immediate\write\@auxout{\string\citation{#2}}\fi%
  \def\@citea{}\@cite{\@for\@citeb:=#2\do%
    {\@citea\def\@citea{, }\@ifundefined
       {b@\@citeb}{{\bf ?}%
       \@warning{Citation `\@citeb' on page \thepage \space undefined}}%
{\csname b@\@citeb\endcsname}}}{#1}}%

%
\def\@citex[#1]#2{%
\if@filesw\immediate\write\@auxout{\string\citation{#2}}\fi%
  \def\@citea{}\@cite{\@for\@citeb:=#2\do%
    {\@citea\def\@citea{; }\@ifundefined
       {b@\@citeb}{{\bf ?}%
       \@warning{Citation `\@citeb' on page \thepage \space undefined}}%
{\csname b@\@citeb\endcsname}}}{#1}}%

%
\def\@biblabel#1{}
\makeatother



\newdimen\bibindent
\bibindent=0.0em
\def\thebibliography#1{\section*{\refname}\list
   {}{\settowidth\labelwidth{[#1]}
   \leftmargin\parindent
   \itemindent -\parindent
   \listparindent \itemindent
   \itemsep 0pt
   \parsep 0pt}
   \def\newblock{}
   \sloppy
   \sfcode`\.=1000\relax}


\setlength{\baselineskip}{12.7pt}

\title{SUPERVISED MACHINE LEARNING FOR EFFECTIVE MISSILE LAUNCH BASED ON BEYOND VISUAL RANGE AIR COMBAT SIMULATIONS}

\author{Joao P. A. Dantas\\ 
        Andre N. Costa\\
        Felipe L. L. Medeiros\\ 
        Diego Geraldo\\[12pt]
Decision Support Systems Subdivision\\
Institute for Advanced Studies \\
Trevo Cel Av Jose A. A. do Amarante, 1, Putim\\
Sao Jose dos Campos, SP 12228-001, BRAZIL\\
\and
Marcos R. O. A. Maximo\\[12pt]
Autonomous Computational Systems Lab (LAB-SCA)\\
Computer Science Division\\
Aeronautics Institute of Technology\\
Praca Marechal Eduardo Gomes, 50, Vila das Acacias\\
Sao Jose dos Campos, SP 12228-900, BRAZIL\\
\and
Takashi Yoneyama\\[12pt]
Electronic Engineering Division\\
Aeronautics Institute of Technology\\
Praca Marechal Eduardo Gomes, 50, Vila das Acacias\\
Sao Jose dos Campos, SP 12228-900, BRAZIL\\
}

\maketitle

\section*{ABSTRACT}

This work compares supervised machine learning methods using reliable data from constructive simulations to estimate the most effective moment for launching missiles during air combat. We employed resampling techniques to improve the predictive model, analyzing accuracy, precision, recall, and f1-score. Indeed, we could identify the remarkable performance of the models based on decision trees and the significant sensitivity of other algorithms to resampling techniques. The models with the best f1-score brought values of $0.379$ and $0.465$ without and with the resampling technique, respectively, which is an increase of $22.69\%$. Thus, if desirable, resampling techniques can improve the model's recall and f1-score with a slight decline in accuracy and precision. Therefore, through data obtained through constructive simulations, it is possible to develop decision support tools based on machine learning models, which may improve the flight quality in BVR air combat, increasing the effectiveness of offensive missions to hit a particular target. \href{https://github.com/jpadantas/effective-missile-launch}{[Code]}\footnote{ \href{https://github.com/jpadantas/effective-missile-launch}{Code: https://github.com/jpadantas/effective-missile-launch}}  

\section{INTRODUCTION}
\label{sec:1}
In a Beyond Visual Range (BVR) air combat, since pilots do not have visual contact with their opponents, the targets are detected using sensors, such as radar or Radar Warning Receiver (RWR). Lately, due to technological advances in sensors and weapons, BVR air combat has become one of the fundamental elements to achieving air superiority~\shortcite{higby2005promise}. Constructive computer simulations have been widely used to emulate the most diverse BVR air combat situations to assess the effects of new combat tactics, sensors, and armaments, at a reduced cost compared to live exercises~\shortcite{dantas2022asa,costa2019}. In a constructive simulation, an aircraft is modeled as an autonomous agent, which is a computational entity that makes decisions based on data obtained from the environment in which it operates and interacts with other agents~\shortcite{russell2020artificial}. One of constructive BVR simulation's main challenges is mimicking the complex behaviors in all combat phases. A pilot can perform decision-making processes, such as adapting to new combat situations and conducting collective tactics with other aircraft~\shortcite{costa2021formation}. Some research addressed the application of artificial intelligence, game theory, and heuristics in modeling decision-making for autonomous agents in different phases of simulated BVR air combat.

Concerning threat assessment and target selection, some of the available methods found in the literature are: Bayesian optimization algorithm in~\shortciteN{fu2021air}; backpropagation neural network in~\shortciteN{yao2021study}; combination of a genetic algorithm with deep learning in~\shortciteN{li2020multi}; zero-sum game in~\shortciteN{ma2019cooperative}; deep neural network in~\shortciteN{dantas2021lars}; Case-Based Behavior Recognition proposed in~\shortciteN{borck2015case}; and an algorithm that combines fuzzy logic and Bayesian network in~\shortciteN{rao2011situation}. 

In the phase of selection and execution of tactical maneuvers, there are also many approaches in the literature: deep reinforcement learning in~\shortciteN{hu2021application}; Bayesian networks in~\shortciteN{Du2010}; zero-sum differential game in~\shortciteN{garcia2021beyond}; genetic algorithm in~\shortciteN{kuroswiski2020modelagem}; reinforcement learning, artificial neural network and Markov decision processes in~\shortciteN{piao2020beyond}; Hierarchical Multi-Objective Evolutionary Algorithm in~\shortciteN{yang2020evasive}; decision algorithm based on the minimax method in~\shortciteN{kang2019beyond}; Markov decision processes, reinforcement learning, a multilayer perceptron (MLP) neural network, and simulated annealing in~\shortciteN{weilin2018decision}; reasoning based on objectives in~\shortciteN{floyd2017goal}; finite state machines, rule-based scripts, and reinforcement learning in~\shortciteN{toubman2016rapid}; and grammatical evolution and behavior trees in~\shortciteN{yao2015adaptive}.

Our work addresses the decision-making process of estimating the most effective moment for firing a missile in a BVR air combat in a specific scenario. The most decisive moment of firing is the one that guarantees the elimination of the target. 

In~\shortciteN{ha2018stochastic}, the authors modeled the BVR air combat between two aircraft swarms as a zero-sum stochastic game. The firing moment estimation process was designed as a probabilistic function of the target's evasive ability, the missile's speed on the final approach, and the accuracy of the target's information to guide the missile. However, the methodology proposed has not been tested in simulations with higher fidelity concerning real BVR air combat. Also, significant simplifications were made in the process of modeling a BVR air combat: all aircraft in the same swarm fire at the same time, the missiles can pursue targets without the need for support from the radar aircraft that launched them, and the evasive ability of each aircraft is an input parameter.

In~\shortciteN{dantas2018}~and~\shortciteN{dantas2022machine}, the authors designed a MLP neural network using data from constructive simulations to be employed in an embedded device to enhance the pilot's situational awareness in the in-flight decision-making process. Although obtaining promising results, only a single machine learning method was employed to analyze a reduced number of air combat scenarios simulation runs with only one aircraft on each team. \shortciteN{lima2021decision} adopted a similar methodology but used data collected in real air combat exercises with computationally simulated shots.

The main contribution of our work is to compare the application of different machine learning methods to estimate the most effective moment for launching missiles during a BVR air combat, based on data from constructive simulations, which are run through a commercial off-the-shelf framework. Since running the simulations is computationally demanding, the machine learning methods can streamline the predictions regarding missile success to be used in real-time applications. Furthermore, it was observed that, during the simulations, the missiles could not hit their targets in most of the scenarios due to challenging shooting conditions within our experiment design, which led to an imbalanced dataset. Therefore, we employed resampling techniques to improve the predictive model, analyzing accuracy, precision, recall, and f1-score. To the best of our knowledge, this is the first work to address the imbalance in the missile launch results within air combat simulation .

The rest of this paper is organized as follows. The definition of a simulated BVR air combat scenario is presented in Section~\ref{sec:2}. Section~\ref{sec:3} describes the methodology proposed to solve the problem of estimating the best moment for firing missiles using machine learning models. The analysis of the results obtained with the application of the machine learning models is presented in Section~\ref{sec:4}. Finally, Section~\ref{sec:5} describes the conclusions of this work and shares ideas for future work.

\section{SCENARIO DESCRIPTION}
\label{sec:2}

The scenario is formed by two forces that engage each other in a tropical area. The blue side performs a Combat Air Patrol (CAP), classified as a Defensive Counter Air mission. In contrast, the red side takes part in a fighter sweep (Offensive Counter Air) to neutralize the blue patrol, opening space for other types of tasks to be performed later, such as an air-to-ground attack. The red side always employs a 4-aircraft squadron formed by multi-role fighters, while the blue side can have either a 4- or a 6-aircraft squadron, which may also vary between two fighter concepts (defined as type 1 and 2). The concepts represent different classes of fighters; type 1 is a smaller single-engine aircraft (fourth-generation), while type 2 is a larger double-engine stealth aircraft (fifth-generation). These entities were defined within a framework for composable simulations called FLAMES~\shortcite{flames}, which provides several aspects of constructive simulation development and use, including customizable scenario creation, execution, visualization, and analysis, as well as interfaces to constructive, virtual, and live systems.

We modeled the BVR air combat behaviors of the aircraft, i.e., autonomous agents, through behavior trees structured within states of a finite state machine~\shortcite{colledanchise2018behavior}. These behaviors include CAP, target selection, engagement, evasion, and missile firing. Also, the aircraft is equipped with several subsystems: controllers, trackers, datalinks, fusers, radars, RWR, antennas, and weapon systems. We will focus on the ones directly related to the variables used in our experimental design.

The red side employs a squadron formed by four twin-engine, mid-size fifth-generation fighters. These aircraft have internal weapons bays carrying up to 4 BVR air-to-air missiles (BVRAAMs). The blue side can have 4- and 6-aircraft squadrons of either type 1 or 2 aircraft. Type 1 is a light aircraft with a reduced cost, whereas type 2 is much larger. Both are single-engine multi-role fighters with stealth capabilities. The red and blue aircraft's main parameters can be found in Table~\ref{tab:1}.

\begin{table}[htb]
\scriptsize
\setlength{\tabcolsep}{4.75pt}
\setlength{\tabcolsep}{5pt}
\centering
\caption{Main characteristics of the red and blue (type 1 and 2) aircraft.}
\label{tab:1}
{\begin{tabular}{lccclccc}
\hline
Feature & Red & Blue 1 & Blue 2 & Feature & Red & Blue 1 & Blue 2\\ 
\hline
Wing area               & 49.5 m$^2$    & 36 m$^2$      & 43 m$^2$    & Mission fuel weight     & 6000 kg     & 3070 kg       & 5200 kg     \\
Wing span               & 11.52 m       & 8.9 m         & 9.5 m       & Maximum weight          & 26205 kg    & 13300 kg      & 22214 kg    \\
Wing incidence          & 1$^\circ$     & 0$^\circ$     & 1$^\circ$   & Dry weight              & 17805 kg    & 9085 kg       & 14638 kg    \\
Wing 25\% sweep         & 27$^\circ$    & 38.4$^\circ$  & 42$^\circ$  & Maximum payload         & 8250 kg     & 6000 kg       & 8000 kg     \\
Wing dihedral           & -4$^\circ$    & 0$^\circ$     & -2$^\circ$  & Radius of action        & 648 NM      & 1000 NM       & 700 NM      \\
Wing twist              & 2.5$^\circ$   & 2$^\circ$     & 2.5$^\circ$ & Cruise altitude         & 11000 m     & 10668 m       & 11000 m     \\
Length fuselage         & 17.28 m       & 15 m          & 18.96 m     & Cruise Mach             & 0.8         & 0.9           & 0.8         \\
Aspect ratio            & 2.68          & 2.2           & 2.1         & Max Mach (sea level)    & 1.1         & 1.1           & 1.1         \\
Taper ratio             & 0.21          & 0.2           & 0.2         & Thrust dry              & 59038 N     & 64874 N       & 109642 N    \\
Ultimate load factor    & 13.5          & 14.5          & 13.5        & Thrust aferburner       & 91781 N     & 98022 N       & 173550 N    \\
Number of missiles      & 4             & 3             & 6           & Front RCS               & -10 dBm$^2$ & -10 dBm$^2$   & -25 dBm$^2$ \\
\hline
\end{tabular}}
\end{table}

\begin{figure}[htb]
\centering
\subfloat[Initial setting.\label{fig:1a}]{\includegraphics[width=0.49\textwidth]{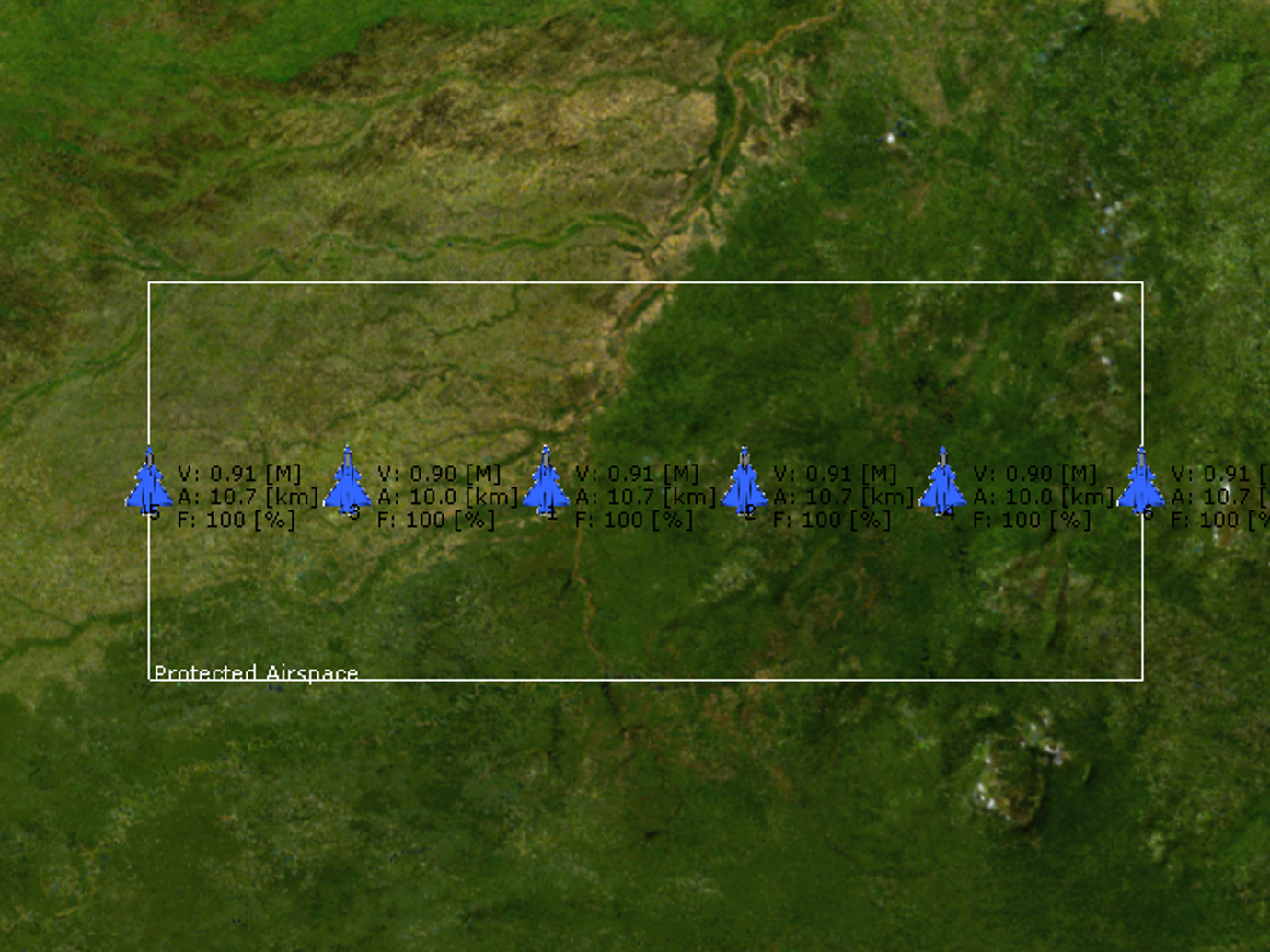}}
\subfloat[Intermediate frame.\label{fig:1b}]{\includegraphics[width=0.49\textwidth]{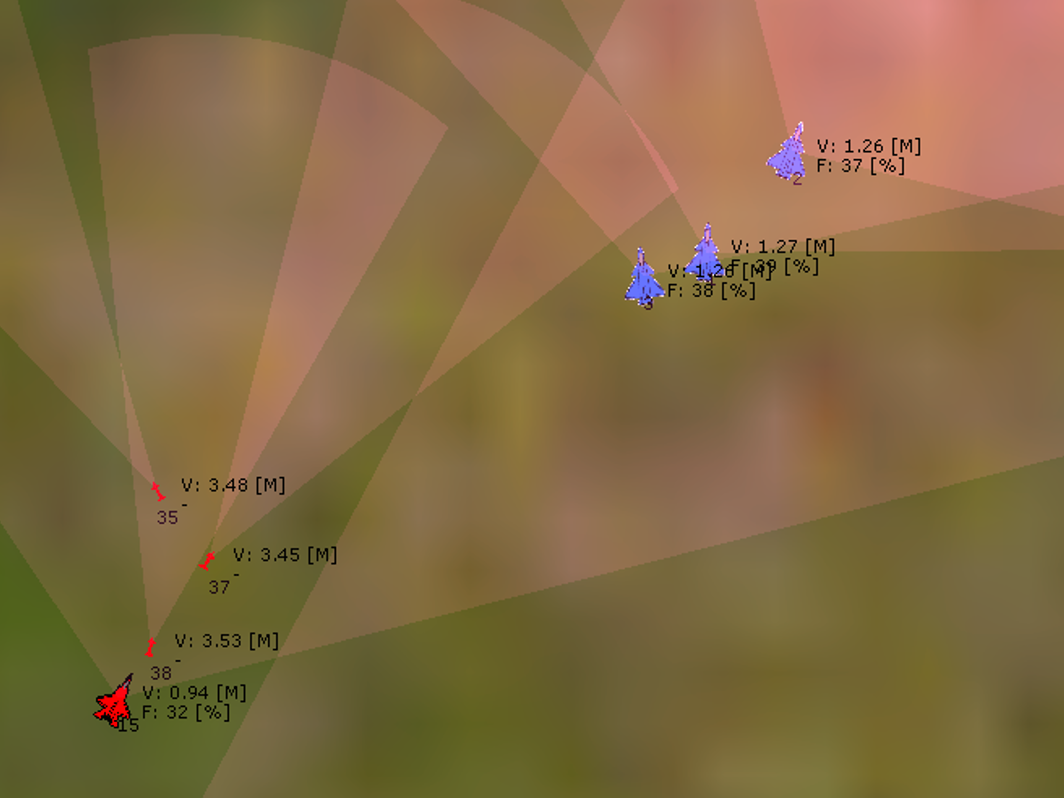}}
\caption{Scenario depictions with (a) 6 blue aircraft in wall formation within protected airspace and (b) radar and missile representations during the simulation execution.}
\end{figure}

Even though each aircraft type had its own baseline front Radar Cross Section (RCS), an additive factor in dBm² was included throughout the design of experiments (DoE) so that different RCSs could be evaluated. Even though just the frontal value was provided, the RCS for each aircraft was determined by a table with values at each angle. Therefore, we applied the additive factor to all values within these tables. These aircraft types are both equipped with BVRAAM missiles with similar characteristics to those employed by the red side. However, different from what was modeled for the red missiles, the activation distance of the blue missiles varied to assess whether this would influence the shot's effectiveness. Additionally, during the different experiment cases, we changed the initial formation of the blue and red aircraft (distance between their CAP points, as in Figure~\ref{fig:1a}). Moreover, in the case of blue aircraft, we modified the radar ranges and the shot philosophy within their behavior trees. We altered flying altitudes and speeds according to the DoE. Figure~\ref{fig:1b} shows a representation of the radars and missiles employed during a simulation execution.

The engagement distances for the aircraft's missiles are dynamically calculated based on factors such as the launcher's altitude and speed and the target's altitude, speed, and azimuth. Therefore, the aircraft's artificial pilot may decide whether to fire or not. The shot philosophy is defined as a percentage of the Weapon Engagement Zone (WEZ), calculated by the earlier parameters, stored on a table. For the red aircraft, this percentage is fixed at $60\%$. The WEZ, in simple terms, represents the range of a given weapon to be employed in a particular shooting stance~\shortcite{dantasbracis2021}.

Besides this percentage, the behavior considers other factors to choose to fire, such as whether the target is already engaged by another friendly aircraft. Additionally, it defines defensive maneuvers and any other action performed by the aircraft. The data that feed the behavior trees come mainly from the aircraft's sensors. The most important ones are the RWR and the radar, which provide the unit's tracklist, i.e., the combined list containing all its possible targets.

The radar could operate in two distinct modes: scan and track. In the scan mode, the radar can cover larger areas with a smaller range. This is used to detect the aircraft within the theater of operations. Diversely, the track mode focuses on a particular target (or set of targets) to follow its movements and prepare to fire. The track mode range is usually larger than the scan mode, which was therefore considered $60\%$ of the track mode range. While tracking, the radar power is concentrated on more specific regions, providing more extensive ranges at the expense of situational awareness concerning the surrounding areas.

\section{METHODOLOGY}
\label{sec:3}

We model the estimation of the most effective moment for firing missiles in BVR air combat simulations as a classification problem, a classic machine learning challenge. This problem consists of classifying a set of instances of features, independent variables, into two classes defined by a single dependent variable: value $1$ indicating an effective shot; and value $0$ meaning an ineffective shot. We employed some of the most relevant supervised machine learning methods to estimate the most effective moment for firing missiles in BVR air combat simulations: Logistic Regression (LR), K-Nearest Neighbors (KNN), Support Vector Machines (SVM), Artificial Neural Networks (ANN), Naive Bayes (NB), Random Forest (RF), and Extreme Gradient Boosting (XGBoost); to the interested reader, we refer to \shortciteN{geron2019hands}. Concerning the resampling techniques, we applied oversampling methods such as Synthetic Minority Oversampling Technique (SMOTE)~\shortcite{chawla2002smote} and Adaptive Synthetic Sampling Approach (ADASYN)~\shortcite{he2008adasyn}. Also, we introduced undersampling methods such as Tomek Links (TL)~\shortcite{Tomek1976} and Edited Nearest Neighbor (ENN)~\shortcite{Wilson1972}. Besides, we analyzed the datasets using hybrid techniques that use oversampling and undersampling together: SMOTE with Tomek Links (SMOTE-TL)~\shortcite{Batista2004} and SMOTE with Edited Nearest Neighbor (SMOTE-ENN)~\shortcite{Batista2004}. 

We proposed the following steps to solve this problem: design of a set of BVR air combat experiments (Subsection~\ref{subsec:3.1}); acquisition of the shooting dataset, i.e., the training dataset, collected by performing the set of simulations (Subsection~\ref{subsec:3.2}); specifying the variables used in the predictive models (Subsection~\ref{subsec:3.3}); hyperparameters tuning process of all supervised machine learning models applied to the problem (Subsection~\ref{subsec:3.4}); preprocessing of the training dataset (Subsection~\ref{subsec:3.5}); and training and evaluation process of the machine learning models (Subsection~\ref{subsec:3.6}).

\subsection{Design of Experiments}
\label{subsec:3.1}
Considering these scenarios and behavior characteristics, we create an experimental design using the Latin Hypercube Sampling (LHS)~\shortcite{McKay1979}. This method is an alternative to using factorial strategies, presenting a solution for filling the sample space in a better fashion than what is done by a purely random process, such as the Monte Carlo sampling~\shortcite{husslage2011space}. LHS divides the space into a prespecified number of sections from which it randomly samples one point. The variables in Table~\ref{tab:2} form the sample space in this work, and each has a predefined interval to design the samples properly. This sampling method allows the user to specify the number of sample points desirable. In our work, we selected $240$ cases as simulation inputs, each executed $30$ times with different random seeds, which mainly interfered with missile probability of kill (set as $90\%$ for both sides) and behavior evaluation delays since both were functions of these seeds. This probability value means that if the missile can reach a distance of at least $10$ meters from the target, it has a $90\%$ chance of destroying the target.

\begin{table}[htb]
\scriptsize
\setlength{\tabcolsep}{4.75pt}
\centering
\caption{Model input and output variables.}
\label{tab:2}
\begin{tabular}{lccc}
\hline
Description & Min & Max & Unit \\ 
\hline
Blue maneuvering altitude                               & 27.5  & 42.5  & kft       \\
Red maneuvering altitude                                & 27.5  & 42.5  & kft       \\
Blue maneuvering speed                                  & 0.9   & 1.5   & Mach      \\
Red maneuvering speed                                   & 0.9   & 1.5   & Mach      \\
Blue radar track range                                  & 150   & 300   & km        \\
Blue missile range multiplicative factor                & 1     & 2     & -         \\
Blue RCS                                                & -10   & 10    & dBm$^2$   \\
Blue missile activation distance                        & 15    & 30    & km        \\
Blue shot philosophy (in percentage of WEZ)             & 50    & 70    & \%        \\
Initial distance between blue aircraft (in longitude)   & 0.1   & 1.0   & $^\circ$  \\
Initial distance between red aircraft (in longitude)    & 0.1   & 1.0   & $^\circ$  \\
Blue CAP speed                                          & 0.7   & 0.75  & Mach      \\
Red CAP speed                                           & 0.7   & 0.75  & Mach      \\
Blue aircraft concept (type)                            & 1     & 2     & -         \\
Whether there are six blue aircraft (0) or not (1)      & 0     & 1     & -         \\
\hline
\end{tabular}
\end{table}

\subsection{Simulation Results}
\label{subsec:3.2}

With the $240$ cases defined by the LHS, $7{,}200$ simulations were carried out, generating the data reports for analysis. Each run stopped after all aircraft had been destroyed or a simulation time of $30$ minutes had been reached, which is a regular time of BVR air combat engagements. Since the simulations ran faster than in real-time, these $30$ minutes took only an average of $5$ seconds to be run, totaling $10$ hours for the whole experiment. The first dataset generated concerns the number of surviving aircraft after each run. It also contained the total number of missiles fired by the design case but no data concerning the firing conditions. In the second dataset, on the other hand, the focus was primarily on the firing conditions, presenting the positions of the shooter and the target at the moment of launch and the distance and the off-boresight angle between them (line of sight). Since the models used for the simulation are still restricted, the resulting datasets are also subject to restraints that preclude us from sharing them.

\subsection{Variables}
\label{subsec:3.3}

Using mutual information for selecting features in supervised models~\shortcite{Battiti1994}, we chose $12$ variables from the $28$ initial variables in the dataset after considering their distribution, correlation analysis, and operational importance as input and output variables of the proposed models. Table~\ref{tab:3} describes the $11$ features and $1$ target variable used in this work's supervised machine learning models, considering the moment that the shooter aircraft launches the missile towards the target aircraft.

\begin{table}[htb]
\scriptsize
\setlength{\tabcolsep}{4.75pt}
\centering
\caption{Description of the model's input and output variables.} 
\label{tab:3}
\begin{tabular}{lcc} 
\hline
Variables & Unit & Description \\ 
\hline
\texttt{radar\_track\_range}    & km       & Radar track range                                      \\
\texttt{distance}               & km       & Distance between aircraft                              \\
\texttt{missile\_act\_dist}     & km       & Missile activation distance                            \\
\texttt{delta\_altitude}        & m        & Difference between aircraft altitudes                  \\
\texttt{delta\_speed}           & kt       & Difference between aircraft speeds                     \\
\texttt{missile\_range}         & -        & Missile range multiplicative factor                    \\
\texttt{rcs}	                & m$^2$    & Radar cross section                                    \\
\texttt{firerange}              & \%       & Percentage of WEZ                                      \\
\texttt{angle\_uni\_to\_tgt}    & $^\circ$ & Off-boresight angle between aircraft                   \\
\texttt{delta\_heading}         & $^\circ$ & Difference between aircraft headings                   \\
\texttt{concept}                & -        & Which blue aircraft concept type was employed (1 or 2) \\
\texttt{kill}                   & -        & Whether the target aircraft was killed by the missile  \\
\hline
\end{tabular}
\end{table}

\subsection{Hypeparameters Tuning}
\label{subsec:3.4}

We created all the models using the Scikit-learn library, except the XGBoost model that has its library~\shortcite{chen2016xgboost}, and, unless otherwise stated, with its default settings for their hyperparameters~\shortcite{scikit2011}. We perform the grid search algorithm as the technique for tunning all models since we want to use our experience from previous works to find the best set of values, even though it is well-known that random search may avoid the drawbacks of regular intervals~\shortcite{bergstra2012random}.

The LR model is using the default Scikit-learn hyperparameters except for $\texttt{C}=100$, which is the inverse of regularization strength, after performing a grid search exploring in $\{10^{-3}, 10^{-2}, 10^{-1}, 10^{0}, 10^{1}, 10^{2}, 10^{3}\}$.

Regarding the KNN model, we chose $12$ for its hyperparameter \texttt{n\_neighbors}, which is the number of neighbors, after checking different values (from $1$ to $50$) and keeping track of the error rate for each of these models to find the minimum value.

Concerning the SVM model, we searched for \texttt{C}, which defines the amount of violation of the margin allowed, in $\{0.1, 1, 10, 100, 1000\}$ and \texttt{gamma}, a parameter that must be specified to the learning algorithm, in $\{1, 0.1, 0.01, 0.001, 0.0001\}$, resulting in $\texttt{C} = 10$ and $\texttt{gamma} = 0.1$ as the best set.

In the ANN model, we explored \texttt{learning\_rate\_init}, the initial learning rate, in $\{0.001, 0.01, 0.1\}$, \texttt{activation}, representing the activation function for the hidden layer, in $\{\texttt{logistic}, \texttt{tanh}, \texttt{relu}\}$, \texttt{solver}, which is the solver for weight optimization, in $\{\texttt{sgd}, \texttt{adam}\}$, and \texttt{alpha}, the strength of the L2 regularization term, in $\{0.0001, 0.001\}$ . After that, we found $\texttt{learning\_rate\_init} = 0.001$, $\texttt{activation} = \texttt{relu}$, $\texttt{solver} = \texttt{adam}$ and $\texttt{alpha} = \texttt{0.001}$ as the best set.

The NB model is operating the default except for $\texttt{var\_smoothing}=0.002$, which represents the part of the most significant variance of all features that is added to variances for calculation stability, found after running the grid search trying $100$ different values spaced evenly on a log scale from $10^{0}$ to $10^{-9}$.

In the RF model, we searched for \texttt{max\_features}, the number of features to consider when looking for the best split in the model, in  $\{2, 3, 4, 5, 6, 7, 8, 9, 10\}$, \texttt{min\_samples\_leaf}, the minimum number of samples required to be at a leaf node, in $\{3, 5, 8\}$, and \texttt{min\_samples\_split}, the minimum number of samples required to split an internal node, in $\{4, 8, 12\}$. The best hyperparameters set found was $\texttt{max\_features} = 5$, $\texttt{min\_samples\_leaf} = 8$, and $\texttt{min\_samples\_split} = 4$.

In the XGBoost model, we inspect the best hyperparameters between \texttt{gamma}, the minimum loss reduction required to make a further partition on a leaf node, in $\{0.5, 1, 1.5\}$, \texttt{subsample}, the subsample ratio of the training instances, in $\{0.6, 0.8\}$, \texttt{colsample\_bytree}, the subsample ratio of columns when constructing each tree, in $\{0.6, 0.8, 1.0\}$, and \texttt{max\_depth}, the maximum depth of a tree, in $\{3, 4, 5\}$. The best set was $\texttt{gamma} = 1.5$, $\texttt{subsample} = 0.8$, $\texttt{colsample\_bytree} = 0.8$ and $\texttt{max\_depth} = 5$.

After finding each model's best set of hyperparameters, we applied all of the resampling techniques in the dataset using the Imbalanced-learn library~\shortcite{JMLR:v18:16-365} with its default hyperparameters.

\subsection{Preprocessing}
\label{subsec:3.5}

The main goal of standardizing features is to help the convergence of the technique used for optimization~\shortcite{laurent2016}, which we perform to use the algorithms SVM, ANN, KNN, and NB. Data scaling was employed to equally distribute the importance of each input in the learning process~\shortcite{priddy2005artificial}. Standardization is not required for LR, RF, and XGBoost models. Besides, we performed Exploratory Data Analysis (EDA) to identify the general data behaviors. The methods employed in this analysis were correlation and descriptive statistics analysis. 

The effectiveness of the training phase of machine learning models can deteriorate if the training dataset is imbalanced, i.e., if there is not an adequate number of samples relating the values of the features for each of the classes (target values). Through the EDA, we verified this problem, as there were many more samples of the occurrence of shots belonging to class \texttt{NO KILL}, indicating the non-elimination of the target. Thus, we used resampling methods to increase the number of class \texttt{KILL} samples.

\subsection{Models Training and Evaluation Process}
\label{subsec:3.6}

Before performing any training process, we conducted a train-validate-test split, randomly separating the data, allocating $85\%$ for training and validation using a 5-fold cross-validation technique and $15\%$ for testing. The test dataset will allow the evaluation of the machine learning models later. We try every combination of values from the grid search proposed, calculating the performance metric f1-score using 5-fold cross-validation. The grid point that best fits the dataset is the optimal combination of the hyperparameters. After running all the algorithms with the imbalanced training-validate dataset and evaluating it, we perform resampling techniques to improve the number of samples of the minority class. We again employed all the supervised learning algorithms using their best set of hyperparameters with these new balanced datasets. As the model deals with a classification problem, we evaluated the model by observing the metrics accuracy, precision, recall, and f1-score~\shortcite{geron2019hands}. The metrics results were acquired by analyzing the models' predictions and actual values in the test dataset.

\section{RESULTS AND ANALYSIS}
\label{sec:4}

This section examines the exploratory data analysis, the test dataset metrics, and time inference of the proposed supervised machine learning algorithms with and without resampling techniques.

\subsection{Exploratory Data Analysis}
\label{subsec:4.1}

Table~\ref{tab:4} shows an overview of the descriptive statistics of these model's inputs numerical variables. All the model's input variables follow a uniform distribution since these variables were sampled using the LHS.

\begin{table*}[htb]
\scriptsize
\setlength{\tabcolsep}{4.75pt}
\centering
\caption{Descriptive statistics of the model's input numerical variables.}
\label{tab:4}
\setlength{\tabcolsep}{3pt}
\begin{tabular}{lccccccc}
\hline
Variables & Mean & Std & Min & 25\% & Median &      75\% & Max\\ 
\hline
\texttt{radar\_track\_range}    & 217,375.7     & 40,219.6  & 150,289.2     & 182,961.3     & 214,907.9     & 251,175.8     & 285,788.2 \\
\texttt{distance}               & 40,841.9      & 16,529.9  & 3,023.6       & 29,319.6      & 38,078.3      & 50,553.9      & 100,717.4 \\
\texttt{missile\_act\_dist}     & 22,442.6      & 4,353.8	& 15,197.3      & 18,742.0      & 22,735.8      & 26,224.3      & 29,828.1  \\
\texttt{delta\_altitude}        & 768.6        & 2,357.6   & -14,168.3     & -655.3       & 641.8        & 2,284.2       & 13,573.3  \\
\texttt{delta\_speed}           & 12.2         & 71.7     & -232.4       & -38.0        & 16.6         & 63.1         & 263.3    \\
\texttt{missile\_range}         & 1.5          & 0.3      & 1.0          & 1.2          & 1.5          & 1.8          & 2.0      \\
\texttt{rcs}	                & -0.1         & 5.8      & -9.9         & -5.3         & -0.1         & 5.0          & 9.7      \\
\texttt{firerange}              & 60.2         & 5.7      & 50.2         & 55.4         & 60.6         & 65.2         & 69.9     \\
\texttt{angle\_uni\_to\_tgt}    & -0.8         & 21.8     & -44.9        & -9.0         & 0.0          & 4.6          & 44.9     \\
\texttt{delta\_heading}         & 181.9        & 77.1     & -42.8        & 140.5        & 180.0        & 220.0        & 401.5    \\
\hline
\end{tabular}
\end{table*}

Remember that one of the features (\texttt{concept}) and the target variable (\texttt{kill}) are categorical. The distribution of the input variable \texttt{concept} is almost equal, presenting $67{,}238$ ($43.78\%$) samples as \texttt{Type 1} and $86{,}368$ ($56.22\%$) samples as \texttt{Type 2}. On the other hand, the output variable \texttt{kill} is very imbalanced, presenting only $18{,}397$ ($11.98\%$) samples as \texttt{KILL} and $135{,}209$ ($88.02\%$) samples as \texttt{NO KILL}. This imbalance is due to the fact that many of the $240$ simulation cases have generated challenging shooting conditions, which degraded the missile envelope. These conditions could be aircraft with large altitude and speed differences or low aircraft RCS coupled with large missile activation distances, among others.

\begin{figure*}[htb]
\centering
\includegraphics[width=0.70\textwidth]{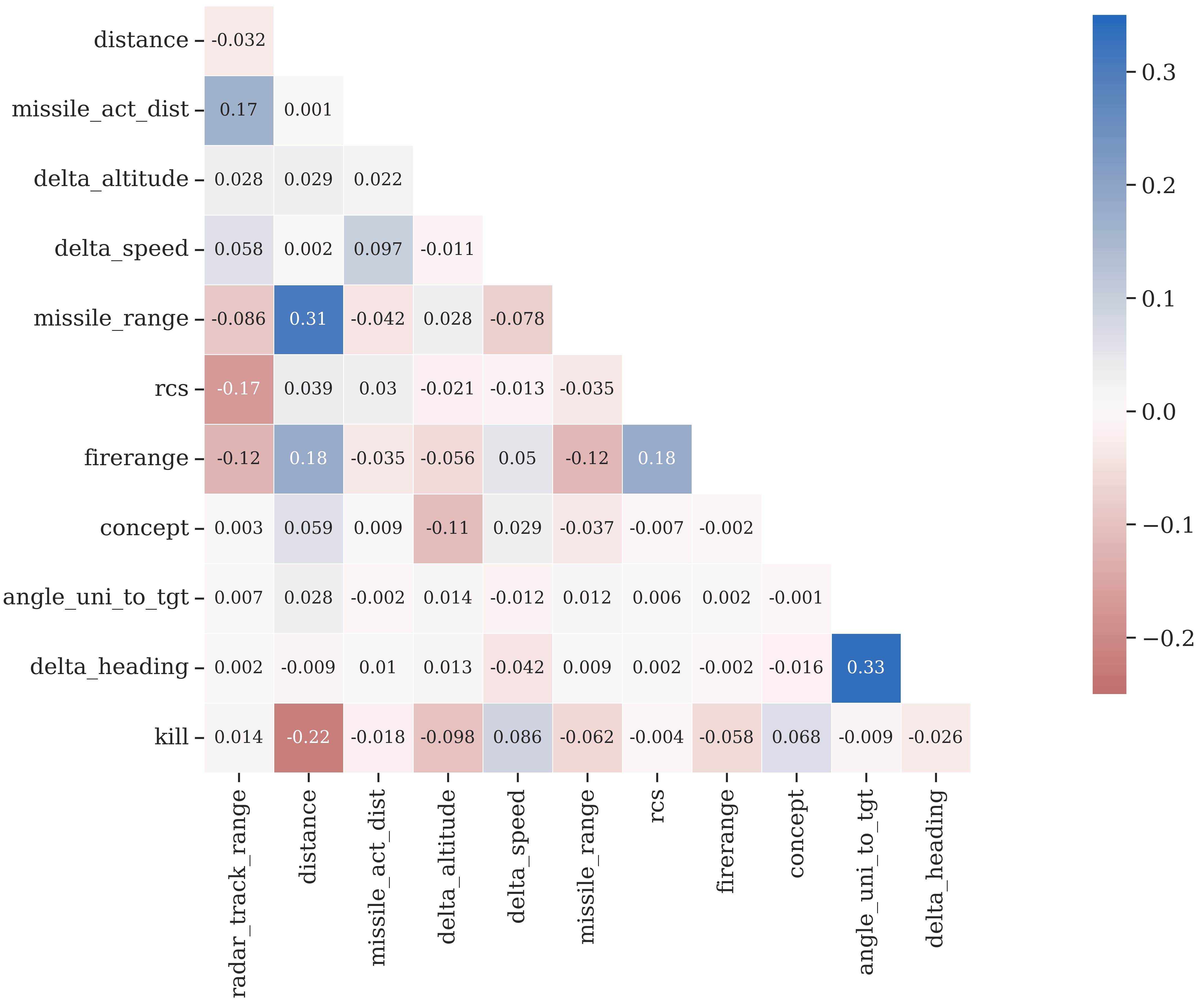}
\caption{Pearson correlation matrix of all model variables.}
\label{fig:2}
\end{figure*}

Pearson's correlation analysis of the variables can be seen in the correlation matrix represented in Figure~\ref{fig:2}. Notice that none of the model's features has a strong correlation with each other, with the largest absolute value being only $0.33$ between \texttt{delta\_heading} and \texttt{angle\_uni\_to\_tgt}, and $0.31$ between \texttt{missile\_range} and \texttt{distance}. The algorithm's performance may deteriorate if two or more variables are tightly related, called multicollinearity~\shortcite{dantasbracis2021}. We may also be interested in the correlation between input variables with the output variable (\texttt{kill}) to provide insight into which variables may or may not be relevant as input for developing a model. Only the variable \texttt{distance} has a slight correlation ($-0.22$) with the target variable. The missile launch analysis is a complex non-linear problem that depends on multiple variables with a low correlation that corroborates the application of supervised machine learning algorithms since they may relate to all these variables.

\subsection{Models Metrics}
\label{subsec:4.2}

Table~\ref{tab:5} shows the metrics, accuracy (ACC), precision (PREC), recall (REC) and F1-score (F1), and the inference time (IT) in milliseconds (ms) obtained after evaluating the test dataset on the supervised machine learning models using or not resampling techniques. Regarding the accuracy, we expect at least metrics better than $0.880$, which is the proportion of the majority class in the imbalanced dataset. If the model consistently predicted the majority class, the accuracy would be $0.880$. Thus, accuracy may not be the most appropriate metric for evaluating the models~\shortcite{menardi2014training}. However, we notice a slight improvement in all models without resampling, except the LR model, with the RF models showing the best results with a low inference time regarding real-time applications ($160.8$ ms to predict the test dataset), as expected since this technique is quite suitable for imbalanced data~\shortcite{khoshgoftaar2007empirical}. Precision, recall, and f1-score can better assess this problem~\shortcite{Jeni2013}.

In general, the resampling techniques such as oversampling methods or hybrid approaches, using oversampling and undersampling techniques together, increased recall and f1-score with a slight decline in the accuracy and precision. Thus, considering it is an imbalanced dataset, each supervised machine learning model and the corresponding resampling technique must be applied in different situations, depending on the interest of having a greater precision or recall, which depends on the actual application of that model.

Low precision refers to cases where models are erroneously reporting that the missile would succeed if launched, but that does not happen in reality. Regarding costs associated with modern air combat, models with high precision avoid unnecessary missile launches and help maintain the aircraft's capability since there is a limit to the number of missiles to be loaded on the aircraft in a regular air combat operation. 

Low recall refers to cases in which the model suggests that the missile would not hit the target aircraft if launched in a given condition, recommending not fire at that moment, though, in reality, that missile would have hit the target aircraft successfully if launched. As target neutralization is a critical factor in BVR air combat, not launching a missile that would probably hit a particular target is a massive issue that good pilots try to avoid, which shows the importance of having a high recall. The F1-score, the harmonic mean of precision and sensitivity, synthesizes the model's performance more broadly. The model using RF and SMOTE obtained the best f1-score overall with low inference time as well ($108.3$ ms), compared to the worst inference time from the model using KNN and SMOTE ($5194.4$ ms).

Therefore, we observe that the SVM model without resampling techniques presents the best precision results ($0.716$), even though it has one of the worst recall metrics. XGBoost and RF models also offer good precision, $0.648$ and $0.686$ respectively, and low recall, both around $0.250$. Considering the analysis of the recall metric itself, in all the models, ADASYN and SMOTE-ENN brought the best increase in this metric. In ANN models, which showed the highest recall value, the ADASYN technique allowed a $224.30\%$ increase in the recall, going from $0.251$ to $0.814$. In the LR models, the increase was from almost null recall values to values that reached $0.711$ using SMOTE-ENN. Thus, resampling techniques demonstrated the capacity to increase the recall metric, even with a slight decrease in accuracy and precision. 

\begin{table}[htb]
\scriptsize
\setlength{\tabcolsep}{4.75pt}
\centering
\caption{Supervised learning classification models metrics and inference time.}
\label{tab:5}
\begin{tabularx}{\textwidth}{lccccclccccc}
\hline
MODEL & ACC & PREC & REC & F1 & IT[ms] & MODEL & ACC & PREC & REC & F1 & IT[ms]\\ 
\hline
 LR                          & 0.877 & 0.421 & 0.003 & 0.006 & 1.4 & ANN + SMOTE-TL          & 0.757 & 0.300 & 0.744 & 0.428 & 157.0 \\
 LR + SMOTE                  & 0.655 & 0.215 & 0.685 & 0.327 & 1.1 & ANN + SMOTE-ENN             & 0.730 & 0.283 & 0.784 & 0.416 & 35.7\\
 LR + ADASYN                 & 0.646 & 0.212 & 0.694 & 0.325 & 1.3 & NB                          & 0.884 & 0.672 & 0.096 & 0.168 & 4.3\\
 LR + SMOTE-TL               & 0.661 & 0.217 & 0.676 & 0.328 & 1.3 & NB + SMOTE                  & 0.641 & 0.208 & 0.690 & 0.320  & 5.0\\
 LR + SMOTE-ENN              & 0.625 & 0.204 & 0.711 & 0.317 & 1.1 & NB + ADASYN                 & 0.599 & 0.196 & 0.736 & 0.310  & 3.3\\
 KNN                         & 0.889 & 0.639 & 0.208 & 0.314 & 3903.1 & NB + SMOTE-TL               & 0.641 & 0.208 & 0.691 & 0.320  & 3.3\\
 KNN + SMOTE                 & 0.763 & 0.296 & 0.678 & 0.412 & 5194.4 & NB + SMOTE-ENN              & 0.590 & 0.194 & 0.743 & 0.307  & 3.2 \\
 KNN + ADASYN                & 0.730 & 0.272 & 0.721 & 0.395 & 4970.0 & RF                          & 0.895 & 0.686 & 0.262 & \textcolor{blue}{\textbf{0.379}} & \textcolor{red}{\textbf{160.8}}\\
 KNN + SMOTE-TL              & 0.763 & 0.297 & 0.685 & 0.414 & 5090.1 & RF + SMOTE                  & 0.851 & 0.415 & 0.528 & \textcolor{blue}{\textbf{0.465}} & \textcolor{red}{\textbf{108.3}}\\
 KNN + SMOTE-ENN             & 0.725 & 0.273 & 0.750 & 0.400 & 4632.8 & RF + ADASYN                 & 0.844 & 0.401 & 0.551 & 0.464  & 175.8\\          
 SVM                         & 0.891 & 0.716 & 0.185 & 0.294 & 266.5 & RF + SMOTE-TL               & 0.848 & 0.408 & 0.537 & 0.463  & 350.0\\  
 SVM + SMOTE                 & 0.766 & 0.308 & 0.729 & 0.433 & 654.7 & RF + SMOTE-ENN              & 0.809 & 0.351 & 0.664 & 0.459  & 35.2\\
 SVM + ADASYN                & 0.722 & 0.276 & 0.785 & 0.409 & 752.8 & XGBoost                     & 0.892 & 0.648 & 0.255 & 0.366  & 17.1\\
 SVM + SMOTE-TL              & 0.766 & 0.307 & 0.727 & 0.432 & 614.5 & XGBoost + SMOTE             & 0.826 & 0.368 & 0.593 & 0.454  & 6.6\\
 SVM + SMOTE-ENN             & 0.737 & 0.287 & 0.775 & 0.419 & 321.345 & XGBoost + ADASYN            & 0.811 & 0.347 & 0.615 & 0.444  & 6.4\\
 ANN                         & 0.890 & 0.638 & 0.227 & 0.335 & 116.6 & XGBoost + SMOTE-TL          & 0.825 & 0.369 & 0.609 & 0.459  & 6.2\\
 ANN + SMOTE                 & 0.765 & 0.308 & 0.735 & 0.434 & 94.4 & XGBoost + SMOTE-ENN         & 0.791 & 0.332 & 0.699 & 0.450  & 6.8\\
 ANN + ADASYN                & 0.703 & 0.267 & 0.814 & 0.402 & 163.1\\
\hline
\end{tabularx} 
\end{table}
\section{CONCLUSIONS AND FUTURE WORK}
\label{sec:5}

This work compares the application of supervised machine learning methods, using resampling techniques to improve the predictive model, on data from air combat constructive simulations to improve the effectiveness of missile launching, and demonstrates the differences in the performance of these several different models.

Without resampling, XGboost and RF brought the most consistent results considering the f1-score. Concerning all oversampling methods or hybrid techniques using oversampling and undersampling techniques, it is possible to indicate that these techniques increase recall and f1-score with a slight decline in accuracy and precision. The model with the best performance, considering the f1-score, without using any resampling techniques was RF which brought $0.379$, with an inference time of $160.8$ milliseconds regarding the time to predict the test dataset. After employing SMOTE, the RF model got $0.465$, the best overall f1-score, increasing $22.69\%$. 

Therefore, since obtaining real air combat data is quite problematic, we show through data obtained through constructive simulations that it is possible to develop decision support tools that may improve flight quality in BVR air combat since they are trying to support effective missile launches. We can use these models in the attempt to enhance the missile launching process by unmanned combat aerial vehicles or aid pilots in real air combat scenarios, increasing the effectiveness of offensive missions.

For future work, we suggest analyzing different scenarios with multiple aircraft equipped with other missiles, which can bring less deterministic simulations. Also, we recommend studying not just the missile launching moment but a sequence of several timeframes to understand better the coordination of the future events of a specific air combat scenario. Besides, since the hyperparameter space for the classifiers is not complete due to computational cost, we propose expanding the grid search space for all supervised machine learning models or, at least, for the models that delivered better results. Finally, we recommend developing more suitable design sampling strategies that allow the choice of other input variables with much more simulation runs to improve the dataset to be analyzed, which can bring more reliability and generalization to the results to enhance the pilot's in-flight situational awareness.

\section*{ACKNOWLEDGMENTS}

This work was supported by Finep (Reference nº 2824/20). Prof~Dr.~Takashi~Yoneyama is partially funded by CNPq -- National Research Council of Brazil through the grant 304134/2-18-0.

\footnotesize

\bibliographystyle{wsc}

\bibliography{demobib}

\section*{AUTHOR BIOGRAPHIES}

\noindent {\bf JOAO~P.~A.~DANTAS} received his B.Sc. degree in Mechanical-Aeronautical Engineering (2015) and his M.Sc. degree in Electronic and Computer Engineering from the Aeronautics Institute of Technology (ITA), Brazil. Currently, he is pursuing his Ph.D. degree at ITA and working as a researcher for the Brazilian Air Force at the Institute for Advanced Studies and in the Robotics Institute at Carnegie Mellon University. His e-mail address is \email{dantasjpad@fab.mil.br}.\\

\noindent {\bf ANDRE~N.~COSTA} received his B.Sc. degree in Mechanical-Aeronautical Engineering (2014) and his M.Sc. degree from the Graduate Program in Electronic and Computer Engineering (2019) from the Aeronautics Institute of Technology, Brazil. Currently, he is a researcher for the Brazilian Air Force at the Institute for Advanced Studies with research interests in modeling and simulation, artificial intelligence, and machine learning. His email address is \email{negraoanc@fab.mil.br}.\\

\noindent {\bf FELIPE~L.~L.~MEDEIROS} received the bachelor’s degree in Computer Science (1999) from the Federal University of Ouro Preto, Brazil, and the master’s (2002) and Ph.D. (2012) degrees in Applied Computing from the National Institute for Space Research, Brazil. He has been working in the field of artificial intelligence with an emphasis on metaheuristics, autonomous agents and simulation. His email address 
is \email{felipefllm@fab.mil.br}.\\

\noindent {\bf DIEGO~GERALDO} was commissioned in 2005 as a graduate of the Brazilian Air Force (FAB) Academy and a fighter pilot in 2006. He received his M.Sc. degree from the Graduate Program in Aeronautical and Mechanical Engineering at the Aeronautics Institute of Technology in 2012. Since then he has been a researcher for FAB at the Institute for Advanced Studies, focusing on modeling and simulation, and artificial intelligence. His email address is \email{diegodg@fab.mil.br}.\\

\noindent {\bf MARCOS~R.~O.~A.~MAXIMO} received the B.Sc. degree (Hons.) (summa cum laude) in Computer Engineering (2012), and the M.Sc. (2015) and Ph.D. (2017) degrees in Electronic and Computer engineering from the Aeronautics Institute of Technology (ITA), Brazil. He is currently a Professor at ITA, being a member of the Autonomous Computational Systems Laboratory (LAB-SCA) with research interests in robotics and artificial intelligence. His email address is \email{mmaximo@ita.br}.\\

\noindent {\bf TAKASHI~YONEYAMA} received the bachelor's degree in Electronic Engineering from the Aeronautics Institute of Technology (ITA), Brazil, in 1975, the M.D. degree in Medicine from the Taubate University, Brazil, in 1993, and the Ph.D. degree in Electrical Engineering from the Imperial College London, U.K., in 1983. He is a Professor of Control Theory with the Department of Electronic at ITA. His research is focused on stochastic optimal control theory. His email address is \email{takashi@ita.br}.

\end{document}